\documentclass{article}

\usepackage{arxiv}

\usepackage[utf8]{inputenc} 
\usepackage[T1]{fontenc}    
\usepackage{hyperref}       
\usepackage{url}            
\usepackage{booktabs}       
\usepackage{amsfonts}       
\usepackage{nicefrac}       
\usepackage{microtype}      
\usepackage{lipsum}		
\usepackage{graphicx}
\usepackage{natbib}
\usepackage{doi}
\usepackage{amsmath}
\usepackage{subcaption}
\usepackage{graphicx}

\title{Unveiling the Training Dynamics of ReLU
Networks through a Linear Lens}

\author{ Longqing Ye
}

\renewcommand{\shorttitle}{\textit{arXiv} Template}

\hypersetup{
pdftitle={A template for the arxiv style},
pdfsubject={q-bio.NC, q-bio.QM},
pdfauthor={David S.~Hippocampus, Elias D.~Striatum},
pdfkeywords={First keyword, Second keyword, More},
}

\begin{document}
\maketitle

\begin{abstract}
Deep neural networks, particularly those employing Rectified Linear Units (ReLU), are often perceived as complex, high-dimensional, non-linear systems. This complexity poses a significant challenge to understanding their internal learning mechanisms. In this work, we propose a novel analytical framework that recasts a multi-layer ReLU network into an equivalent single-layer linear model with input-dependent "effective weights". For any given input sample, the activation pattern of ReLU units creates a unique computational path, effectively zeroing out a subset of weights in the network. By composing the active weights across all layers, we can derive an effective weight matrix, $W_{\text{eff}}(x)$, that maps the input directly to the output for that specific sample. We posit that the evolution of these effective weights reveals fundamental principles of representation learning. Our work demonstrates that as training progresses, the effective weights corresponding to samples from the same class converge, while those from different classes diverge. By tracking the trajectories of these sample-wise effective weights, we provide a new lens through which to interpret the formation of class-specific decision boundaries and the emergence of semantic representations within the network.
\end{abstract}

\section{Introduction}

The remarkable success of deep neural networks (DNNs) across various domains is well-documented \cite{vgg,googlenet,resnet}. However, their "black box" nature remains a major obstacle to their broader application in safety-critical systems and for scientific discovery. A central question in deep learning theory is how these networks learn meaningful representations from data throughout the training process. While significant research has focused on analyzing the distribution of learned weights or the activation patterns of individual neurons, the holistic transformation applied to each specific input sample is often obscured by the network's depth and non-linearity.

The non-linearity in many state-of-the-art networks is introduced by simple, piecewise-linear activation functions, with the Rectified Linear Unit (ReLU) being the most prominent. A ReLU network partitions the input space into a vast number of linear regions, and within each region, the network behaves as a simple affine transformation. Inspired by this, we propose a shift in perspective: instead of viewing the network as a fixed non-linear function, we interpret it as a system that generates a unique linear transformation for each input sample.

Our core concept is the "effective weight" matrix, denoted as $W_{\text{eff}}(x)$. This matrix represents the equivalent linear transformation that a multi-layer, bias-free ReLU network applies to a specific input $x$. The existence of such a matrix is guaranteed by the nature of ReLU, which acts as a data-driven switch, pruning connections for each input and thus defining a sample-specific linear subgraph.

This paper investigates the training dynamics of ReLU networks by analyzing the evolution of these effective weights. We posit two primary hypotheses:
\begin{enumerate}
    \item \textbf{Intra-class Convergence:} For two samples, $x_i$ and $x_j$, belonging to the same class, their corresponding effective weights, $W_{\text{eff}}(x_i)$ and $W_{\text{eff}}(x_j)$, will become more similar as the network learns to generalize.
    \item \textbf{Inter-class Divergence:} For samples belonging to different classes, their effective weights will diverge, reflecting the network's process of carving out distinct decision boundaries in the feature space.
\end{enumerate}
By tracking these per-sample transformations, we aim to provide a more granular and interpretable account of how neural networks learn to classify data, moving beyond static analyses to a dynamic and input-dependent understanding.

\section{Methods}

\subsection{Formulation of the Effective Weight Matrix}

Let us consider a standard L-layer feed-forward neural network without bias terms, where the transformation at each layer $l$ is defined by $h_l = \sigma(W_l h_{l-1})$, with $h_0 = x$ being the input vector and $\sigma(\cdot)$ being the ReLU activation function. The ReLU function can be expressed as $\sigma(z) = \max(0, z)$.

This activation function can be rewritten using a diagonal matrix $D_l(x)$ whose diagonal entries are defined by the activation status of the neurons at layer $l$ for the input $x$:
\begin{equation}
    D_l(x)_{ii} = 
    \begin{cases} 
      1 & \text{if } (W_l h_{l-1})_i > 0 \\
      0 & \text{if } (W_l h_{l-1})_i \leq 0 
    \end{cases}
\end{equation}
Using this notation, the output of layer $l$ can be written as a linear transformation of the previous layer's output:
\begin{equation}
    h_l = D_l(x) W_l h_{l-1}
\end{equation}
Note that the matrix $D_l(x)$ is dependent on the input $x$ as it is determined by the forward propagation of $x$ up to layer $l-1$.

By recursively expanding this expression from the output layer back to the input layer, we can express the network's final output (pre-softmax), $f(x)$, as a single linear transformation of the input $x$:
\begin{equation}
    f(x) = (D_L(x) W_L) \dots (D_2(x) W_2) (D_1(x) W_1) x
\end{equation}
From this, we define the input-dependent effective weight matrix $W_{\text{eff}}(x)$ as the product of all layer-wise, data-activated transformation matrices:
\begin{equation}
    W_{\text{eff}}(x) = \prod_{l=L}^{1} (D_l(x) W_l)
\end{equation}
This matrix $W_{\text{eff}}(x)$ has the same dimensions as a single-layer linear model's weight matrix and encapsulates the complete, albeit localized, transformation that the deep network applies to the input $x$.

\section{Experimental Design for Tracking Weight Evolution}

To empirically investigate the dynamics of the effective weights, we designed an experiment using a controlled architecture and dataset.

\subsection{Model, Data, and Training}
We employ a simple yet effective model architecture consisting of a three-layer Multi-Layer Perceptron (MLP) as a backbone network, followed by a linear classification head. Crucially, the layers within the MLP backbone are implemented without bias terms to adhere to the formulation in Equation (4). The model is trained on the MNIST dataset, a standard benchmark for image classification. Training is performed using a standard optimization algorithm (Adam) with a cross-entropy loss function.

\subsection{Protocol for Tracking and Visualization}
To capture the evolution of the network's learned representations, we establish a tracking protocol that monitors the effective weights at discrete points during training.
\begin{enumerate}
    \item \textbf{Sampling:} A fixed subset of the test dataset, constituting 10\% of the total test samples, is selected. Using test data allows us to evaluate the network's generalization behavior.
    \item \textbf{Snapshotting:} At initialization (before training begins) and then periodically after a fixed number of training iterations, we pause the training process.
    \item \textbf{Computation:} For each sample $x_i$ in the selected subset, we compute its corresponding effective weight matrix $W_{\text{eff}}^{(t)}(x_i)$ from the MLP backbone at that specific training snapshot $t$.
    \item \textbf{Dimensionality Reduction:} The space of effective weight matrices is high-dimensional. To visualize their structure, each matrix $W_{\text{eff}}^{(t)}(x_i)$ is flattened into a high-dimensional vector, $\text{vec}(W_{\text{eff}}^{(t)}(x_i))$. We then apply t-Distributed Stochastic Neighbor Embedding (t-SNE) to this collection of vectors to project them into a 2D space for visualization.
\end{enumerate}

\section{Experimental Results and Analysis}

\begin{figure}[h!]
    \centering
    \begin{subfigure}[b]{0.49\textwidth}
        \centering
        \includegraphics[width=\textwidth]{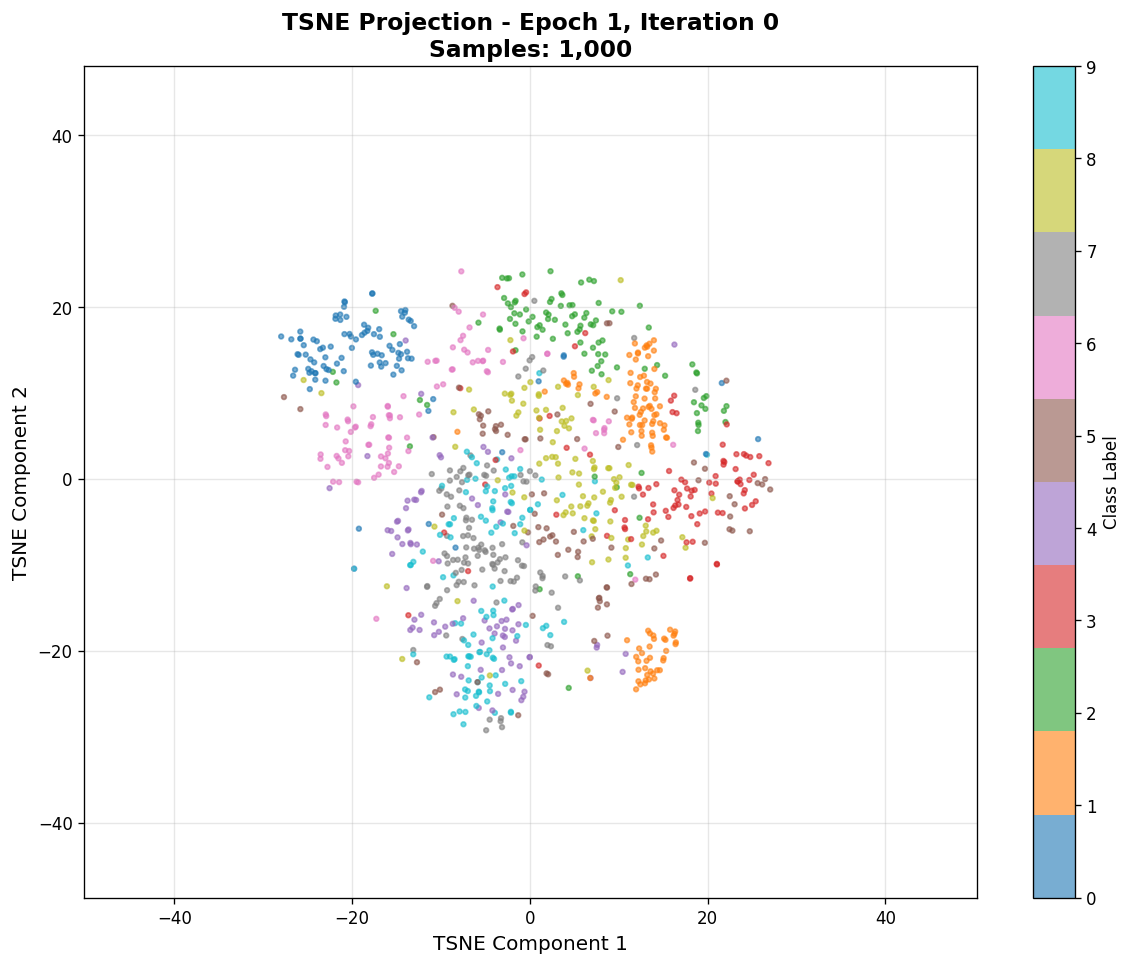}
        \caption{Initial state (Epoch 1, Iteration 0)}
        \label{fig:initial_state}
    \end{subfigure}
    \hfill
    \begin{subfigure}[b]{0.49\textwidth}
        \centering
        \includegraphics[width=\textwidth]{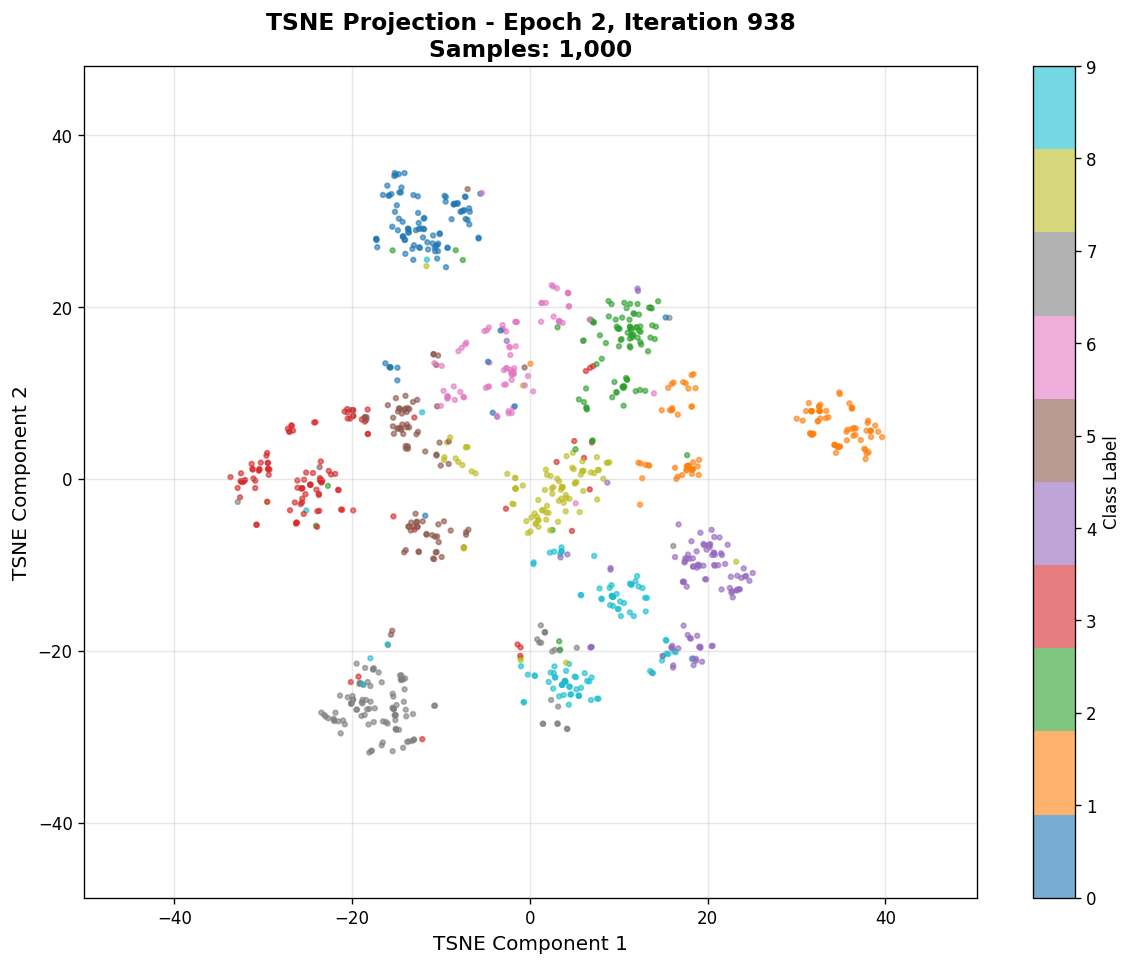}
        \caption{Trained state (Epoch 2, Iteration 938)}
        \label{fig:trained_state}
    \end{subfigure}
    \caption{t-SNE visualization of the effective weight vectors $W_{\text{eff}}(x)$ for 1,000 MNIST test samples. Each point represents the effective weight for a single sample, colored by its true class label. (a) Initial state. (b) After training.}
    \label{fig:tsne_evolution}
\end{figure}

The t-SNE projections of the effective weight manifold, presented in Figure \ref{fig:tsne_evolution}, offer a direct visualization of the network's representation learning process. By contrasting the initial and final states, we can trace the evolution from a partially structured, class-confused space to a highly organized, semantically effective manifold.

\subsection{The Manifold's Evolution: From Initial Biases to Learned Structure}

The initial state of the manifold (Figure \ref{fig:initial_state}) provides a revealing snapshot of the network's pre-learning biases. The distribution is not uniformly chaotic. Instead, two key phenomena are apparent:
\begin{enumerate}
    \item \textbf{Early Separability and Confusion:} Some visually distinct classes, such as '0' and '6', already show a degree of separation from other classes. Conversely, a large group of visually similar classes, most notably '4', '7', and '9', are heavily intermingled in a dense, central cloud.
    \item \textbf{Intra-Class Fracturing:} Furthermore, classes with high internal variance, such as the digit '1', are partitioned into distinct sub-clusters. This indicates that the untrained network is sensitive to significant stylistic variations, initially treating them as different sub-categories requiring different transformations. This fracturing is a strong indicator of the network's sensitivity to significant stylistic variations. A plausible source for this particular split can be found within the dataset itself: the MNIST dataset is known to contain different stylistic variants of the digit '1'. A primary distinction can be made between a simple vertical stroke, and another common variant that is distinguished by the presence of a flat bottom bar. The untrained network, acting as an unintentional "style detector," would naturally map these visually distinct patterns to different regions in the effective weight space. This suggests the network initially treats high-variance members of the same category as different sub-classes.
\end{enumerate}
This initial state reveals that the network begins not as a blank slate, but with a complex predisposition shaped by the interaction between its random weights and the intrinsic structure of the data.

The final, trained state (Figure \ref{fig:trained_state}) demonstrates the profound impact of the learning process. The manifold has undergone a dramatic reorganization, resulting in a much clearer structure. The primary transformation is the significant increase in inter-class separation and intra-class cohesion. The dense, confused cloud of classes has been untangled, with most classes now occupying their own distinct regions. For instance, the clusters for '4', '7', and '9' are now largely separable. Simultaneously, the clusters for most classes have become far more compact and tightly gathered than in their initial state.

\subsection{Implications: Adaptive Processing and Specialized Transformations}

This detailed evolution reveals the sophisticated strategy of deep non-linear learning. The primary goal of training is to organize the manifold to be effective for classification, which it achieves by separating the major class clusters. The persistence of the class '1' sub-clusters, however, provides a deeper insight.

It suggests that the network has learned that forcing a single, unified transformation for a high-variance class like '1' may be suboptimal. Instead, it maintains specialized processing strategies for the different stylistic variations. It develops what can be considered "conceptual sub-classes" within its transformations to effectively handle the different modes in the data.

The role of training, in this light, is to ensure that these different, specialized transformation strategies all lead to the same correct semantic conclusion (i.e., a '1' classification). The final proximity of the sub-clusters shows that the network has learned their shared identity, while their continued existence shows it has retained specialized handling for their visual differences.

This is a fundamental advantage of deep networks over linear models. A linear model is constrained to a single, one-size-fits-all transformation. In contrast, a deep network acts as an adaptive processor. It possesses the flexibility to develop and maintain a repertoire of nuanced, sub-class-specific effective weights ($W_{\text{eff}}(x)$) to tackle data with high intra-class variance. This ability to learn a mixture of specialized yet semantically unified transformations is a hallmark of its powerful representation learning capability.

\section{Conclusion}

In this work, we have presented a new perspective for interpreting the learning dynamics of multi-layer ReLU networks. By recasting a deep, non-linear architecture into an equivalent single-layer model with input-dependent "effective weights" ($W_{\text{eff}}(x)$), we have established a novel framework for analyzing the transformation that the network applies to each individual sample. This approach shifts the focus from the static, global parameters of the network to the dynamic, local function it computes.

Our experimental results, visualized through t-SNE projections of these effective weights, provided compelling evidence for our central hypothesis. We have demonstrated that the training process orchestrates a remarkable evolution in the space of these effective transformations: from an initial, random state to a highly structured manifold. This final state is characterized by the distinct clustering of effective weights for samples of the same class and clear separation between weights of different classes. This visually confirms that representation learning in this context can be understood as the process of learning a meta-function that generates class-specific linear operators.

The significance of this work is twofold. First, it offers a powerful and intuitive tool for neural network interpretability, allowing us to "watch" the formation of a structured representation space as the network learns. Second, it provides a theoretical bridge, connecting the behavior of complex, deep non-linear systems to the more tractable domain of dynamic linear models.

\section{Limitations and Future Work}

We acknowledge the limitations of the current study, which in turn open up exciting avenues for future research. Our framework was developed for a simplified, bias-free, feed-forward ReLU network. Generalizing this "linear lens" to more complex, modern architectures presents both theoretical challenges and opportunities to uncover deeper principles.

A primary limitation is the exclusion of common architectural components. Extending the framework to include bias terms is a direct next step, which would transform our equivalent linear model into a more general affine one: $f(x) = W_{\text{eff}}(x)x + b_{\text{eff}}(x)$. The interplay between the effective weight and the effective bias in shaping the decision boundaries remains an open question. Similarly, incorporating residual connections, which can be viewed as adding an identity matrix to a block's effective weight—would allow our method to analyze how networks like ResNets learn to control information flow by bypassing transformations for certain inputs.

A more fundamental challenge arises from components that break the element-wise non-linearity assumption. Normalization layers (e.g., Batch or Layer Norm) are a key example. Their output for a single neuron depends on the statistics of a larger set of values, breaking the simple algebraic structure required for an exact effective weight calculation.

This leads to a broader question regarding more complex activation functions (e.g., GELU, Swish). While it may not be possible to derive an exact effective weight for these smooth, non-linear functions, it might be feasible to compute a local linear approximation, such as by using the Jacobian of the activation function at the pre-activation value. This "approximated" effective weight, while computationally expensive, could be a powerful theoretical tool.

\bibliographystyle{unsrtnat}
\bibliography{references}  






\end{document}